\documentclass[10pt,twocolumn,letterpaper]{article}

\usepackage{cvpr}
\usepackage{times}
\usepackage{epsfig}
\usepackage{graphicx}
\usepackage{amsmath}
\usepackage{amssymb}
\usepackage{dsfont}
\usepackage{comment}

\usepackage[pagebackref=true,breaklinks=true,letterpaper=true,colorlinks,bookmarks=false]{hyperref}

\cvprfinalcopy 

\ifcvprfinal\fi
\begin{document}

\title{Bilinear Graph Networks for Visual Question Answering}

\author{Dalu Guo, Chang Xu, Dacheng Tao\\
	UBTECH Sydney AI Centre, School of Computer Science, FEIT, 
	\\University of Sydney, Darlington, NSW 2008, Australia\\
	\{\tt\small dguo8417@uni., c.xu@, dacheng.tao@\}sydney.edu.au
}

\maketitle

\begin{abstract}
This paper revisits the bilinear attention networks in the visual question answering task from a graph perspective. The classical bilinear attention networks build a bilinear attention map to extract the joint representation of words in the question and objects in the image but lack fully exploring the relationship between words for complex reasoning. In contrast, we develop bilinear graph networks to model the context of the joint embeddings of words and objects. Two kinds of graphs are investigated, namely image-graph and question-graph. The image-graph transfers features of the detected objects to their related query words, enabling the output nodes to have both semantic and factual information. The question-graph exchanges information between these output nodes from image-graph to amplify the implicit yet important relationship between objects. These two kinds of graphs cooperate with each other, and thus our resulting model can model the relationship and dependency between objects, which leads to the realization of multi-step reasoning. Experimental results on the VQA v2.0 validation dataset demonstrate the ability of our method to handle the complex questions. On the test-std set, our best single model achieves state-of-the-art performance, boosting the overall accuracy to 72.41\%. 
\end{abstract}
\vspace{-10pt}

\section{Introduction}
The developments in computer vision and natural language processing enable the machine to deal with complicated tasks that require the integration and understanding of vision and language, e.g. image captioning \cite{anderson2018bottom}, visual grounding \cite{yu2018rethinking, fukui2016multimodal}, visual question answering (VQA) \cite{antol2015vqa, gao2019dynamic, yu2019deep}, and visual dialog \cite{das2017visual, guo2019image}. Compared with image captioning that is to simply describe the topic of an image, VQA needs a complex reasoning process to infer the right answer for a variety of questions. Visual grounding aims to locate the related objects in the image, but VQA takes a further step to convert this information into human language. In addition, VQA is the basic and vital component in visual dialog. Considering the challenges and significance of VQA, increasing research attention has been attracted to it. 

Given an input image and a question, representative VQA models, e.g. Stacked Attention Networks (SAN) \cite{yang2016stacked}, Multimodal Compact Bilinear Pooling (MCB) \cite{fukui2016multimodal}, and Multimodal Low-rank Bilinear Attention Networks (MLB) \cite{kim2016hadamard}, first generate grid image features by ResNet \cite{he2016deep} and represent the question as the last hidden state of Long Short-Term Memory (LSTM) \cite{hochreiter1997long}, and then attend to the image features based on the question vector to ground the target objects; the question vector and the weighted image features are finally projected into a unified embedding for answer prediction. Bilinear Attention Networks (BAN) \cite{kim2018bilinear} notice that these methods neglect the interaction between words in the question and objects in the image and propose to build a bilinear co-attention map considering each pair of multi-modal channels. Furthermore, Dynamic Fusion with Intra- and Inter-modality (DFAF) \cite{gao2019dynamic} and Deep Modular Co-Attention Networks (MCAN) \cite{yu2019deep} consider intra-attention within each modality and inter-attention across different modalities by the scaled dot-product attention from Transformer \cite{vaswani2017attention}. 

However, BAN lacks comprehensive exploitation of the interactions between words in questions for modeling their context. The linear way of using scaled dot-product to calculate the attention within single modality (the queries, keys, and values come from the kind of nodes), such as textual features \cite{vaswani2017attention, devlin2018bert} and visual features \cite{chen2019graph, yang2018graph}, is less expressive to fully capture the complex relationship within the multi-modal inputs.

In this paper, we develop bilinear graph networks for visual question answering. We first investigate the bilinear attention map between words in the question and objects in the image from a new graph perspective, then we highlight the importance of exploiting the intra-modality relationship between words in the question and exploring the cross-modality relationship between the question and image for complex reasoning. Two graphs are established to formulate these two kinds of relationships. The image-graph focuses on exploring visual features of the image to their related textual features for joint embeddings, which links the semantic information of words with factual information of the image. The question-graph exploits information across different joint embeddings in terms of words, which amplifies the implicit yet important relationships between objects. Given these two graphs cooperating with each other, the resulting VQA model is able to reason complex and compositional questions. 

We conduct experiments on VQA v2.0 dataset \cite{goyal2017making}. On the validation dataset, our one-layer graph networks boost the accuracy by 0.62\% compared with BAN, and graphs of multiple layers show advantages on multi-step reasoning for long and complex questions, evidenced by a total 1.4\% improvement. With the help of pre-trained language model, BERT \cite{devlin2018bert}, our graphs gain an extra 1.5\% increase. On the test-std dataset, our model achieves state-of-the-art performance, increasing the overall accuracy to 72.41\%.

\section{Related Work}
In this section, we will first introduce the related research on VQA and then the graph neural networks on both text-based and visual-based tasks.

\textbf{Visual Question Answering (VQA)}: VQA is a task to answer the given question based on the input image. The question is usually embedded into a vector with LSTM \cite{hochreiter1997long}, and the image is represented by the fixed-size grid features extracted from a pre-trained model, such as ResNet \cite{he2016deep}. Then both of these features are combined by addition or concatenation \cite{antol2015vqa} before being projected into a unified vector for answer prediction through a multilayer perceptron (MLP). However, not all features of the image are related to the given question, while some of them should be filtered out before generating the unified vector, therefore attention mechanism is introduced to learn the weight of each grid feature. Stack Attention Networks (SAN) \cite{yang2016stacked} learn the visual attention through multi-steps, trying to answer the question progressively. Dual Attention Networks (DAN) \cite{nam2016dual} learn visual and textual attention respectively via the memory vector. Due to the different distributions of question and image features, the outer product of both features has a better explanation and performance compared with the linear combination. But because of its high dimension output, it is hard to be optimized. Multimodal Compact Bilinear Pooling (MCB) \cite{fukui2016multimodal} is approaching this process by calculating the count sketch of two features and convolving them in Faster Fourier Transform (FFT) space. Nevertheless, MCB uses sampling features instead of the original ones, which leads to bias and needs a large projected dimension to reduce it. Hadamard Product for Low-rank Bilinear Pooling (MLB) \cite{kim2016hadamard} models the common vector with a low-rank matrix by an element-wise multiplication, and Multi-modal Factorized Bilinear Pooling (MFB) \cite{yu2017mfb} increases the rank from 1 to $k$ to accelerate the convergence rate and improve the model's robustness. Furthermore, Bilinear Attention Networks (BAN) \cite{kim2018bilinear} learn the textual and visual attention simultaneously, which builds a mapping from the detected objects of the image to the words of the question.

\textbf{Graph Neural Network (GNN)}: GNN is used to build the relationship between nodes like social network, citation link \cite{hamilton2017inductive}, knowledge graph \cite{kipf2016semi}, protein-protein interaction \cite{velickovic2017graph}, etc.. It overcomes the limitation of Euclidean distance between each node in the inputs and involves more context information from neighbors. In text-based tasks, such as machine translation and sequence tagging, GNN breaks the sequence restriction between each word and learns the graph weight by attention mechanism, such as Transformer \cite{vaswani2017attention}, which makes it easier to model longer sequence than LSTM and Gated Recurrent Units (GRU) \cite{chung2014empirical}, since each node is directly linked with others via learned weights instead of through hidden state and gates. Pre-training of Deep Bidirectional Transformers (BERT) \cite{devlin2018bert}, which is trained on a large corpus with unsupervised learning approaches, can be easily explained and transferred to other tasks. In image-based tasks, GNN gathers information from all the grids \cite{wang2018non, chen2019graph} or proposals \cite{liu2018structure} other than surroundings whose size is limited by the receptive fields of Convolution Neural Networks (CNNs), and it aggregates features over coordinate space to improve the performance of object detection and scene generation \cite{yang2018graph}. Motivated by these models, DFAF \cite{gao2019dynamic} and MCAN \cite{yu2019deep} consider all the relationships between inputs by calculating the attention weight with scaled dot-product, including word and word, object and object, and object and word. Pretraining Task-agnostic Visiolinguistic Representations (Vilbert) \cite{lu2019vilbert} even fine-tune BERT model by reconstructing the image region categories and words as well as predicting the alignment of the image and its caption. 

\section{Preliminaries}
The goal of VQA task is to answer the given question $T$ based on the input image $I$. With the object-detector Faster-RCNN \cite{ren2015faster, anderson2018bottom}, we convert the input image $I$ into object features $V = (v_1, \dots, v_n)$ with $v_i \in R^D$, where $n$ is the number of detected objects, and $D$ is the feature dimension. The question $(t_1, \dots, t_m)$ is a sequence of $m$ words. It can be encoded using either LSTM \cite{hochreiter1997long} or Transformer \cite{vaswani2017attention, devlin2018bert} to $Q = (q_1, \dots, q_m)$, where $Q = \text{LSTM}(T)$ or $Q = \text{Transformer}(T)$, and $Q \in R^{C \times m}$, where $C$ is the dimension of output features. In order to represent the common vector of $v \in V$ and $q \in Q$, a weight matrix $W_i$ is introduced to calculate the scalar output $f_i$ and can be approximated with multiplication of two sub-matrix $U_iV_i^\top$ following MLB \cite{kim2016hadamard} (bias terms are omitted without loss of generality):
\vspace{-5pt}
\begin{equation} \label{eq:mlb}
f_i = q^\top W_i v \approx q^\top U_i V_i^\top v = \mathds{1}^\top (U_i^\top q \circ V_i^ \top v),
\vspace{-5pt}
\end{equation}
where $W_i \in R^{C \times D}$, $U_i \in R^{C \times d}$, $V_i \in R^{D \times d}$, $\mathds{1} \in R^d$ is a vector with all elements equal to 1, and $\circ$ is Hadamard product (element-wise multiplication). This decomposition makes the rank of matrix $W_i$ to be at most $d \le \text{min}(C, D)$. To obtain the out feature $f \in R^K$, two three-dimension tensors, $U \in R^{C \times d \times K}$ and $V \in R^{D \times d \times K}$, are learned, and empirically $d$ is set to 1, resulting in $U \in R^{C \times K}$ and $V \in R^{D \times K}$ for simplicity. 

However, the question features $Q$ and image features $V$ are multiple channels, BAN \cite{kim2018bilinear} reduces both input channels simultaneously and obtains a unified representation of them. It first calculates a bilinear attention map $G \in R^{m \times n}$ between $Q$ and $V$, conditioned on which, it then generates the joint embedding $z$ as follows:
\vspace{-5pt}
\begin{equation} \label{eq:ban1}
z = \mathrm{BAN}(Q, V; G).
\vspace{-5pt}
\end{equation}
The attention map $G$ is defined as:
\vspace{-5pt}
\begin{equation} \label{eq:ban2}
G =\mathrm{softmax}\Big(((\mathds{1} \cdot \mathbf{p}^\top) \circ \sigma(Q^\top \mathbf{U'})) \sigma(\mathbf{V'}^\top V) \Big),
\vspace{-5pt}
\end{equation}
where $\mathbf{U'} \in R^{C \times K'}$, $\mathbf{V'} \in R^{D \times K'}$, $\mathbf{p} \in R^{K'}$ are variables to be learned, $K'$ denotes the shared embedding size, and $\sigma$ is the ReLU activation function denoted as $\sigma(x)=\max(x,0)$. Notice that the softmax function works on the rows and columns, i.e. $\sum_{i=1}^{m}\sum_{j=1}^{n}G_{i,j} = 1$. The logit $G'_{i,j}$, element of $G$ before softmax, is the output of low-rank bilinear pooling as:
\vspace{-5pt}
\begin{equation} \label{eq:hahadamard}
G'_{i,j} = \mathbf{p}^\top (\sigma(\mathbf{U'}^\top q_i) \circ \sigma(\mathbf{V'}^\top v_j)).
\vspace{-5pt}
\end{equation}
The matrix $\mathbf{p}^\top$ projects the unified vector of $q_i$ and $v_j$ into a scalar to represent the relation between them.

Then the $k$-th element value of joint embedding $z \in R^K$ is given by:
\vspace{-5pt}
\begin{equation}
z_k = \sum_{i=1}^{m}\sum_{j=1}^{n}G_{i,j}\sigma(q_i^\top \mathbf{U}_k)\sigma(\mathbf{V}_k^\top v_j),
\vspace{-5pt}
\end{equation}
where $\mathbf{U} \in R^{C \times K}$, $\mathbf{V} \in R^{D \times K}$ are the parameters to be optimized. It can also be rewritten as:
\vspace{-5pt}
\begin{equation} \label{eq:ban3}
z_k = \sigma(Q^\top \mathbf{U})^\top_k G \sigma(V^\top \mathbf{V})_k,
\vspace{-5pt}
\end{equation}
where $(Q^\top \mathbf{U})_k \in R^m$ is the $k$ column of $Q^\top \mathbf{U}$, and $(V^\top \mathbf{V})_k \in R^n$ is the $k$ column of $V^\top \mathbf{V}$. 

After that, we input $z$ to a classifier such as MLP to calculate the score $p_i$ for answer $a_i \in A$ and choose the highest one as the predicted answer, where $A$ is the answer set.

\section{Bilinear Graph Networks} \label{sec:bgn}
\begin{figure*}[t!]
	\centering
	\includegraphics[width=1.0\linewidth]{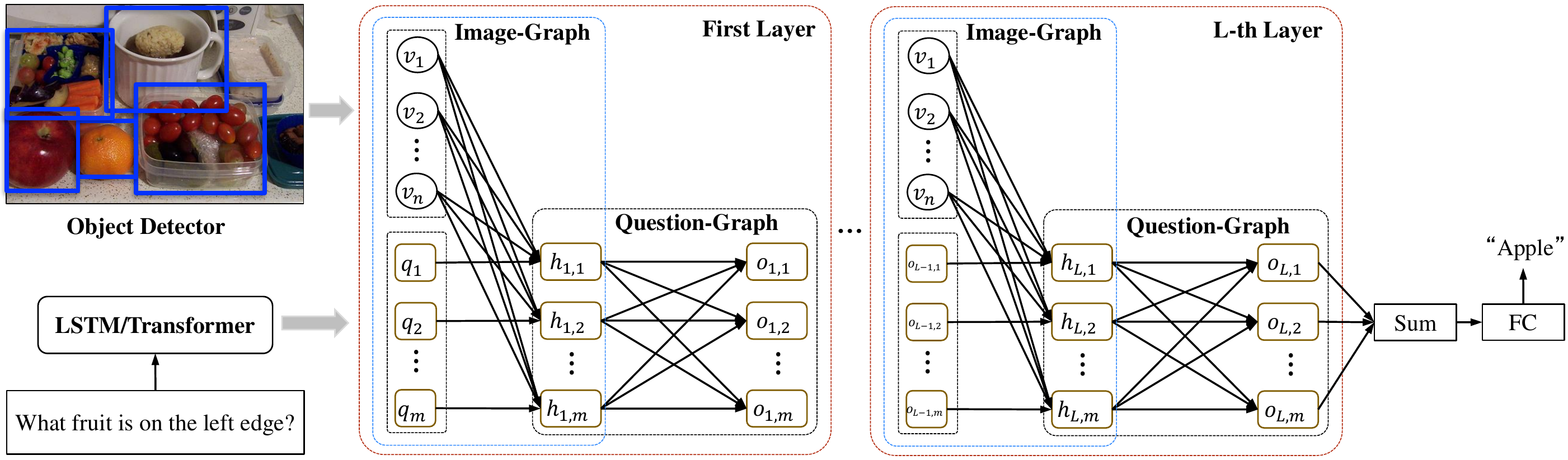}
	\caption{Architecture of our model. The image-graph builds the relationship between words and objects, and the question-graph learns the relationship between joint embeddings in terms of words. The two graphs cooperate with each other to predict the answer.}
	\label{fig:framework}
\vspace{-5pt}
\end{figure*}
The graph attention network and its variant, Transformer, are efficient in modeling the relationship within single modality, such as textual nodes \cite{vaswani2017attention, devlin2018bert}, visual nodes \cite{chen2019graph, yang2018graph}, and citation nodes \cite{velickovic2017graph}, whose outputs can be calculated as:
\vspace{-5pt}
\begin{equation} \label{eq:g1}
\text{Tr}(\mathbb{Q}, \mathbb{K}, \mathbb{V}) = \text{softmax}(\mathbb{Q} \mathbb{K}^\top) \mathbb{V},
\vspace{-5pt}
\end{equation}
where $\mathbb{Q}, \mathbb{K}$, and $\mathbb{V}$ denote the queries, keys, and values respectively, and the softmax function only works on the rows. Motivated by Eq. (\ref{eq:g1}), we can easily illustrate BAN from the perspective of graph.

Given the calculation of $z_k$ in Eq. (\ref{eq:ban3}), Eq. (\ref{eq:ban1}) can be reformulated as:
\vspace{-5pt}
\begin{align} \label{eq:ban4}
Z'^\top &= \mathrm{BGN}(Q, V; G) = \sigma(Q^\top \mathbf{U}) \circ G^a \sigma(V^\top \mathbf{V}), \\ \label{eq:ban5}
z &= Z' G^b,
\vspace{-5pt}
\end{align}
where $Z' \in R^{K \times m}$, $G^a \in R^{m \times n}$, $G^b \in R^m$, $G^b_i=\sum_{j=1}^{n}G_{i,j}$, and $G^a_{i,j} = \frac{G_{i,j}}{G^b_i}$. The output nodes $Z'=(z'_1, \dots, z'_m)$ are calculated based on the input nodes $\{Q \cup V\}$ and their attention weight $G^a$. The attention map $G^a$ in Eq. (\ref{eq:ban4}) is equivalent to graph weight $\text{softmax}(\mathbb{Q} \mathbb{K}^\top)$ in Eq. (\ref{eq:g1}), and $\sigma(V^\top\mathbf{V'})$ is the graph value $\mathbb{V}$. Looking into the definition of attention map in Eq. (\ref{eq:ban2}), the map $G^a$ implies how much information should flow from the nodes $V$ to the nodes $Q$. $(\mathds{1} \cdot \mathbf{p}^\top) \circ \sigma(Q^\top \mathbf{U'})$ and $\sigma(V^\top \mathbf{V'})$ correspond to query $\mathbb{Q}$ and key $\mathbb{K}$ in Eq. (\ref{eq:g1}) respectively. Instead of simply using scaled dot-product, low-rank bilinear pooling is utilized to overcome the different distributions of $Q$ and $V$ based on Eq. (\ref{eq:hahadamard}). Moreover, Eq. (\ref{eq:g1}) only considers single modality of inputs, while VQA models need to consider the multi-modal inputs (i.e. image and question). An additional Hadamard product of $\sigma(Q^\top \mathbf{U})$ and $G^a \sigma(V^\top \mathbf{V})$ is thus included in Eq. (\ref{eq:ban4}) to generate the output nodes $(z'_1, \dots, z'_m)$, where $z'_i \in R^K$. Finally, the joint embedding $z$ represents the whole graph by summarization of all nodes in $Z'$ based on their weight $G^b$ in Eq. (\ref{eq:ban5}). 

Even though Eqs. (\ref{eq:ban4}) and (\ref{eq:ban5}) provide an elegant approach to investigate the relationship between question features $Q$ and image features $V$, a simply summarization over columns of $Z'$ in Eq. (\ref{eq:ban5}) cannot fully address the connections between the joint embeddings $(z'_1, \cdots, z'_m)$ corresponding to words. Given the question and image in Figure \ref{fig:framework}, BAN (i.e. Eqs. (\ref{eq:ban4}) and (\ref{eq:ban5})) can locate a variety of fruits in the image according to the word `fruit' in the question, but it is unaware of the relative position of each fruit from others by mixing all the information (i.e. the summarization in Eq. (\ref{eq:ban5})), thus we want each joint embeddings $z'$ to extract its related information from other items instead of an overall representation. Hence, we are motivated to develop bilinear graph networks, as shown in Figure \ref{fig:framework}, which has two kinds of graphs, i.e. image-graph and question-graph. The image-graph learns to build the relationship between words and objects and generates their joint embeddings, while the question-graph will update the joint embeddings in terms of words by exploiting their interactions. 

We also find that the right answer may not be decided at once, therefore we stack our graphs to make the words interact with the objects as well as words themselves for multiple times.

\textbf{Difference from other graph-based methods.} Though we also investigate the VQA problem from a graph view, our model has several differences from existing graph-based methods. Compared with MUREL \cite{cadene2019murel}, representing the question as a single vector to fuse with the image features at each step and emphasizing the relationship between objects, our method pays attention to modeling the relationship between words and objects as well as between words and words. Regarding DFAF \cite{gao2019dynamic}, MCAN \cite{yu2019deep}, and Vilbert \cite{lu2019vilbert}, all of them use the scaled dot-product \cite{vaswani2017attention} to model the graph weight between image and question as well as linearly combining both features to compute their join embeddings, which is less effective in modeling the representation of multi-modal inputs. We reformulate BAN as a bilinear graph between question and image and reveal its disadvantages, then we propose the image-graph and question-graph to solve it, which has a better explanation.

\subsection{Image-Graph}
\begin{figure}
	\centering
	\includegraphics[width=1.0\linewidth]{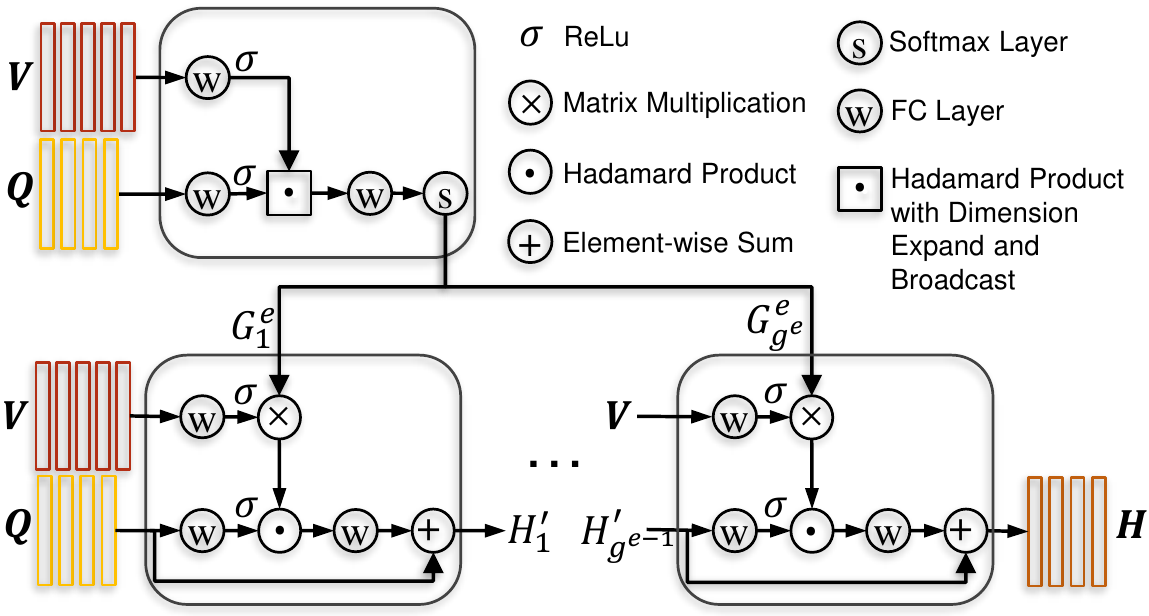}
	\caption{Illustration of multiple glimpses of our image-graph. Each glimpse of graph weight $G^e$ is computed by utilizing the bilinear attention network between Q and V, then the Hadamard product of question features and weighted image features aim to represent their joint embeddings.}
	\label{fig:framework1}
\vspace{-5pt}
\end{figure}
The major target of the image-graph is to locate the objects related to semantic information of each word in the question. Beginning with Eq. (\ref{eq:ban4}), we have a multi-glimpse extension as shown in Figure \ref{fig:framework1}. 

Consider the graph $\mathcal{G} = \{\mathcal{V}, \mathcal{E}\}$, where $\mathcal{V}$ and $\mathcal{E}$ are the set of nodes and edges respectively. The image-graph has $\mathcal{V}=\{Q \cup V\}$ and $\mathcal{E}=G^{e}$, where $Q \in R^{C \times m}$ are textual features of the question and $V \in R^{D \times n}$ are visual features of the detected objects, and $G^{e} $ are the computed graph weights based on $Q$ and $V$. To joint model the graph between image and question from different representation subspaces, we extend $G^{e}$ to multiple glimpses following \cite{kim2018bilinear, vaswani2017attention}, resulting in $G^{e} \in R^{m \times n \times g^e}$, where $g^e$ is the number of glimpse. The $j$-th graph attention is computed as:
\vspace{-5pt}
{\small
\begin{equation} \label{eq:g2}
G_{j}^{e} =\mathrm{softmax}\Big((((\mathds{1} \cdot \mathbf{p}_{j}^{e\top}) \circ \sigma(Q^\top\mathbf{U'}^{e})) \sigma(V^\top \mathbf{V'}^e)^\top \Big),
\end{equation}
}
where the parameters $\mathbf{U'}^e$ and $\mathbf{V'}^e$ are shared among glimpses except for $\mathbf{p}_j^e$, which can be seen from the upper part of Figure \ref{fig:framework1}. After learning the graph attention, we use Eq. (\ref{eq:ban4}) to generate the joint embeddings as:
\vspace{-5pt}
\begin{equation} \label{eq:g3}
\begin{split}
H_j'^\top &= \mathrm{BGN}^e_j(Q, V; G^e_{j}) \\
&= \sigma(Q^\top \mathbf{U}_j^e) \circ G^e_j \sigma(V^\top \mathbf{V}_j^e),
\end{split}
\vspace{-5pt}
\end{equation}
where $H_j' \in R^{K \times m}$ represents the output of image-graph at glimpse $j$.

Instead of concatenation \cite{vaswani2017attention, devlin2018bert, velickovic2017graph} of joint embeddings from each glimpse, we follow BAN to use the residual form to integrate previous learned joint embeddings as shown in the lower part of Figure \ref{fig:framework1}, then Eq. (\ref{eq:g3}) becomes:
\vspace{-5pt}
\begin{equation} \label{eq:g4}
H_j' = W^e_j\mathrm{BGN}^e_j(H_{j-1}', V, G^e_j)^\top + H_{j-1}',
\vspace{-5pt}
\end{equation}
where $H_0' = Q$, and $W^e_j \in R^{C \times K}$ projects the joint embeddings to the same dimension of $Q$. By convention, we use the output of the last glimpse to represent the whole image-graph, denoted as $H = H_{g^e}'$.

\subsection{Question-Graph}
For the question-graph, similarly, we have the graph nodes $\mathcal{V}=H$ and graph weight $\mathcal{E}=G^{r}$, where $H \in R^{C \times m}$ are the output nodes of the image-graph, and $G^{r} \in R^{m \times m \times g^r}$ are the self-attention graph weights of multiple glimpses based on $H$ denoted as:
\vspace{-5pt}
{\small
\begin{equation} \label{eq:g5}
G_{j}^{r} =\mathrm{softmax}\Big((((\mathds{1} \cdot \mathbf{p}_{j}^{r\top}) \circ \sigma(H^\top\mathbf{U'}^{r})) \sigma(H^\top \mathbf{V'}^r)^\top \Big).
\vspace{-5pt}
\end{equation}
}
The structure of our question-graph is similar to the image-graph in Figure \ref{fig:framework1}, except that both inputs are $H$. Different from Eq. (\ref{eq:ban5}), which summarizes the outputs from the image-graph based on $G^b$ to represent the whole graph, $G^r$ in Eq. (\ref{eq:g5}) learns the context of each node for exchanging their information. Based on the graph weight $G^r$, nodes of the question-graph at glimpse $j$ gather information from others and are represented as Eq. (\ref{eq:g4}):
\vspace{-5pt}
\begin{equation} \label{eq:g6}
O_j' = W^r_j\mathrm{BGN}^r_j(O_{j-1}', H; G^r_j)^\top + O_{j-1}',
\vspace{-5pt}
\end{equation}
where $W^r_j \in R^{C \times K}$ and $O_0' = H$. The outputs of question-graph $O$, abbreviated version of $O_{g^r}'$, can be utilized to answer the question by summarizing all the nodes to represent the whole graph.

As we mentioned above, the question may be compositional and complex that needs multi-step reasoning, thus we form the basic module of our bilinear graph networks with one image-graph following by one question-graph, and we stack the module for multiple layers to compose our framework shown in Figure \ref{fig:framework}. The first layer of the image-graph takes textual nodes $Q$ as query to locate the related visual information in $V$ and outputs their joint nodes $H_1$, and the higher layer of it takes the outputs of $i-1$ layer of the question-graph, $O_{i-1}$, as query to involve more visual information related to the prior knowledge. The layer $i$ of question-graph aims at exchanging the information between nodes of $H_i$ to model the context and gets $O_i$ for prior knowledge of image-graph or answer prediction.

After stacking $L$ layers of bilinear graph networks, we summarize all the nodes of $O_{L}$ to represent the whole graph and pass it to a two-layer MLP for classification:
\vspace{-5pt}
\begin{equation}
p = W^{a'}\sigma(W^a O_{L} \cdot \mathds{1}),
\vspace{-5pt}
\end{equation}
where $W^a \in R^{2C \times C}, W^{a'} \in R^{|A|\times 2C}$, and $|A|$ is the size of $A$.

\section{Experiments}
In this section, we evaluate our bilinear graph networks on VQA v2.0 dataset \cite{antol2015vqa, goyal2017making}. We first introduce this dataset and then describe our implementation details and results, and finally the qualitative analysis. 

\subsection{Dataset}
\textbf{VQA v2.0 dataset}: The dataset was built based on the MSCOCO images \cite{lin2014microsoft}, and it contains 1.1M questions asked by human and each question is annotated by ten people. Compared with v1.0 dataset \cite{antol2015vqa}, it emphasizes the visual understanding by reducing the text bias. The dataset is split into three parts: training, validation, and test, which have 80k images and 444 questions, 40k images and 214k questions, and 80k images and 448k questions respectively. The answers of the training and validation dataset are published for training model, while those of the test dataset are unknown and should be predicted by the proposed model before being uploaded to the server for performance evaluation. Based on the answer category, the questions can be classified into three types, i.e. yes/no, number, and others. We train our models with different settings on training dataset and evaluate their accuracy on validation dataset by the tools from \cite{antol2015vqa}, then we pick the settings of the best model for training on the training and validation dataset with extra data from Visual Genome \cite{krishna2017visual} that has 108k images and 1M questions, reporting results on test-server.
	
\subsection{Implementation Details}
We construct the answer vocabulary by restricting to the words that appear in the training and validation dataset more than eight times, resulting in $|A| = 3,129$. We then truncate or pad a question's length $m$ to 15 words, and the weight of padding tokens in question-graph $G^r$ will be set to $-\infty$ before softmax to reduce its impact. There are two methods to encode the question, one is LSTM, and the other one is Transformer. For the former one, we pass the question through a one-layer LSTM, whose input dimension of each word is 600, 300 of which is learned by our model and another 300 from pre-trained GloVe vector \cite{pennington2014glove} is fixed, and the output dimension $C$ is 1,024. For the latter one, we encode the words by summing their corresponding token embeddings and position embeddings and project the outputs of the last layer of Transformer into vectors with dimension $C$ following by $\tanh$, where $\tanh(x)=\frac{e^x-e^{-x}}{e^x+e^{-x}}$. We extract object features from a Faster-RCNN model \cite{anderson2018bottom} pre-trained on Visual Genome, which has 1,600 object classes. For each image, we obtain top $n = 100$ objects based on their probabilities with their object features and regions, and each object feature is presented by mean-pooling of their convolutional features with $D = 2,048$. The joint embedding size $K$ and $K'$ are set to 1,024, and the rank $d$ is set to 3 during computing the graph attention weights in the image-graph and question-graph to increase its capacity. In order to save memory in each layer to make our network go deeper, we reduce the glimpse number from 8 (best performance in BAN) to $g^e = g^r = 4$. Weight Normalization \cite{salimans2016weight} and Dropout \cite{srivastava2014dropout} with $p = 0.2$ are added after each linear mapping to stable the output and prevent from over-fitting. Due to the fact that there might exist multiple correct answers for a question, we utilize the binary cross-entropy loss (BCE) as loss function, which is calculated as:
\vspace{-5pt}
\begin{equation}
L = -\sum_{i = 1}^{|A|}(y_i \log \phi(p_i) + (1-y_i)\log(1 - \phi(p_i))),
\vspace{-5pt}
\end{equation}
where $y_i= \min(\frac{\text{number of people that provided answer} a_i}{3}, 1)$, and $\phi(x)$ is the sigmoid function denoted as $\phi(x)=\frac{1}{1 + e^{-x}}$. Adamax \cite{kingma2014adam}, a variant of Adam, is used to optimize our model. The initial learning rate is 0.001 and grows by 0.001 every epoch until reaching 0.004 for warm start, keeps constant until the eleventh epoch and decays by 1/4 every two epochs to 0.00025. The batch size is 128. 

\subsection{Ablation Study}
We conduct several ablation studies to verify the contribution of each module in our bilinear graph networks (BGNs). The first four lines in Table \ref{table:vqa_model2} show the accuracy of BAN on the VQA v2.0 validation dataset, and BAN-4 and BAN-8 represent the model with 4 and 8 glimpses respectively. It can be seen that simply stacking the module of BAN can improve the accuracy to some extent (0.35\% and 0.46\% in the two-layer and three-layer model respectively) compared with the one-layer model. Although the multi-layer BAN might gain more visual information related to the global representation in Eq. (\ref{eq:ban5}) without exchanging context information, it is not clear about the relationship between entities in question. In contrast, our one-layer model, BGNs $\times$ 1, gains an accuracy 0.62\% and 0.43\% higher than BAN-4 $\times$ 1 and BAN-8 $\times$ 1 respectively, proving the effectiveness of our proposed question-graph even with fewer glimpse. However, if we only stack the question-graph for multiple times (V-graph + Q-graph $\times$ 2 and V-graph + Q-graph $\times$ 3) with only one-layer image-graph, the performance grows slower than that of the BGNs $\times$ 2, this might be caused by that the question-graph can only propagate the information already learned by the image-graph but cannot involve more factual information required in the image to answer the questions. If we replace the proposed bilinear graph network in question-graph with the scaled dot-production (SDP), the accuracy declines (-0.06\% and -0.41\% than BGNs with the same layer) and grows slightly (0.09\%) by stacking the graph. It can be explained that though the inputs of question-graph $H$ are the same type of nodes, the nodes themselves are hybrid and the linear method cannot fully express their relations. By stacking three layers of the BGNs, our model achieves 67.21\% on the overall accuracy, which is chosen as the best model.
\begin{table}
	\centering
	\resizebox{0.6\linewidth}{!}{
	\begin{tabular}{lc}
    	\hline
    	\textbf{Model} & \textbf{VQA Score} \\
    	\hline
    	BAN-4 $\times$ 1 & 65.81 \\
    	BAN-4 $\times$ 2 & 66.16 \\
    	BAN-4 $\times$ 3 & 66.27 \\
    	BAN-8 $\times$ 1 & 66.00 \\
    	\hline
    	BGNs $\times$ 1 & 66.43 \\
    	V-graph + Q-graph $\times$ 2 & 66.69 \\
    	V-graph + Q-graph $\times$ 3 & 66.73 \\
    	\hline
    	(V-graph + SDP) $\times$ 1 & 66.37 \\
    	(V-graph + SDP) $\times$ 2 & 66.46 \\ 
    	BGNs $\times$ 2 & 66.87 \\
    	BGNs $\times$ 3 & \textbf{67.21} \\
    	BGNs $\times$ 4 & 67.06 \\
     	\hline
    \end{tabular}}
\vspace{3pt}
\caption{Score on VQA v2.0 validation dataset. $\times L$ denotes stacking $L$ layers of the proposed model. V-graph is short for image-graph, Q-graph for question-graph, BGNs for (V-graph + Q-graph), and SDP for scaled dot-product.}
\label{table:vqa_model2}
\vspace{-5pt}
\end{table}

\begin{figure}
	\centering
	\includegraphics[width=1.0\linewidth]{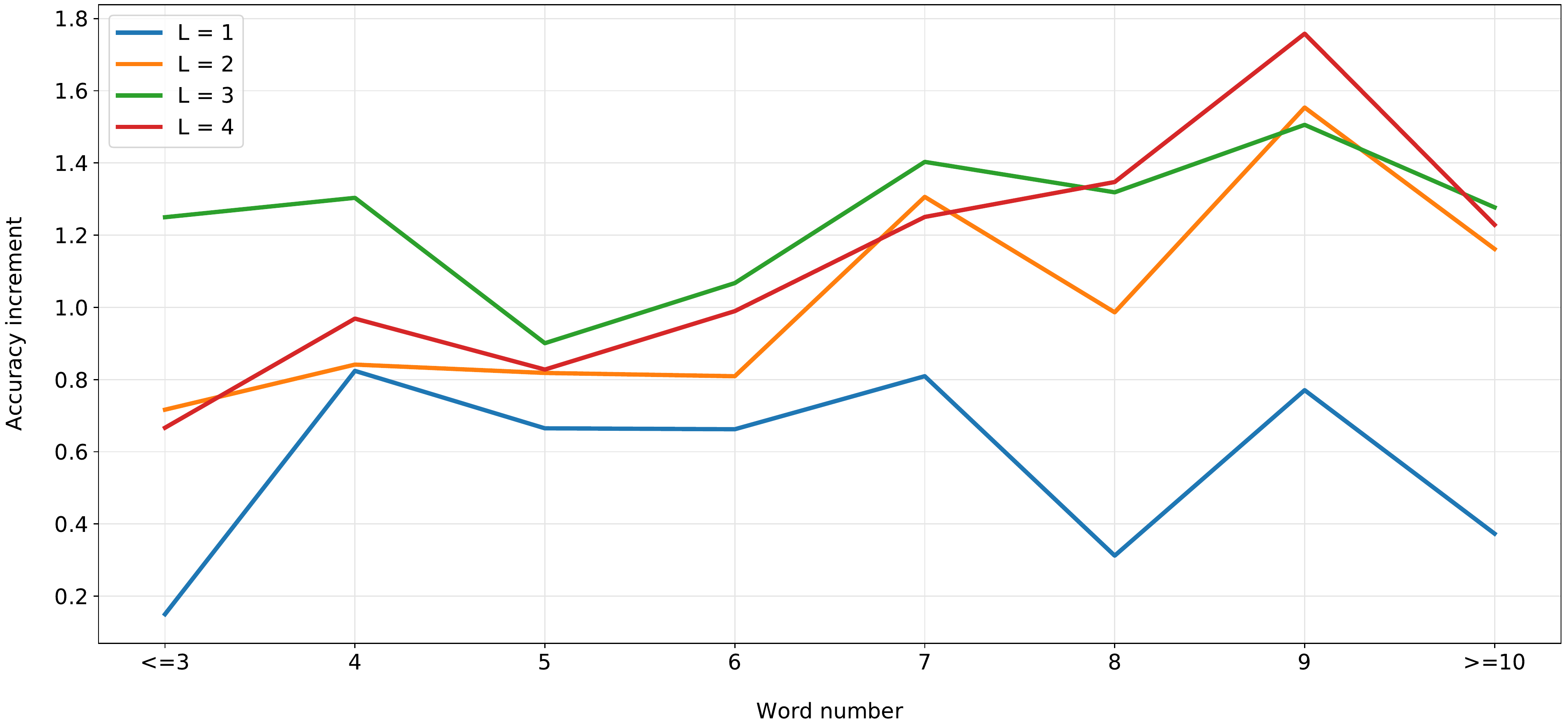}
	\caption{Score increase of our models (with layer=1,2,3,4) compared with one-layer BAN model on VQA v2.0 validation dataset.}
	\label{fig:vqa}
	\vspace{-5pt}
\end{figure}

Additionally, we investigate the absolute increase of the score of our models compared with the single-layer BAN on questions with varied lengths to show the ability of our model on multi-step reasoning in Figure \ref{fig:vqa}. Our models with different layers outperform the one-layer BAN, especially on long questions. The one-layer model does not perform as well as the other three models for long questions due to its shallow graphs. With more layers, our model becomes better at long questions and achieves a 1.7\% increase at word number of nine. What interests us is why the performance drops at four-layer. Comparing the three-layer model and four-layer model, the former one works better in short questions (word number $<$ 8) which take 79\% of all questions, while the latter one has a higher score in long questions, this may explain the performance drop. This phenomenon also inspires us to design a network in the future to classify the questions to fit different layers of graphs.

\begin{table}
	\centering
	\resizebox{0.7\linewidth}{!}{
		\begin{tabular}{lcc}
			\hline
			\textbf{Model} & \textbf{$lr \times$} & \textbf{VQA Score} \\
			\hline
			BGNs $\times$ 1 + LSTM & 1 & 66.43 \\
			\hline
			BGNs $\times$ 1 + Base & 0 & 66.52 \\
			BGNs $\times$ 1 + Base & 0.001 & 67.62 \\
			BGNs $\times$ 1 + Base & 0.01 & 68.09 \\
			BGNs $\times$ 1 + Base & 0.1 & 67.84 \\
			\hline
			BGNs $\times$ 1 + Large & 0.01 & 68.20 \\
			BGNs $\times$ 2 + Large & 0.01 & 68.50 \\
			BGNs $\times$ 3 + Large & 0.01 & \textbf{68.71} \\
			\hline
		\end{tabular}}
		\vspace{3pt}
		\caption{Influence of BERT on our models.}
		\label{table:vqa_model3}
		\vspace{-5pt}
\end{table}
Furthermore, we explore the influence of BERT \cite{devlin2018bert} on our method, since it is trained on large text corpus, therefore it has better generalization and representation of textual features. So we replace the LSTM with it when modeling the question and fine-tuning its weight with different strategies. By using the base model of BERT without fine-tuning (BGNs $\times$ 1 + Base with $lr \times 0$), the accuracy increases slightly, and by increasing its learning rate, it boosts and achieves the best performance at $lr \times 0.01$. With this learning rate, we switch to the large model that is deeper and wider than the base one, the performance grows and keeps going by stacking our bilinear graph model on it, proving that our model is effective and compatible with BERT.

\begin{figure*}
	\centering
	\includegraphics[width=0.30\linewidth]{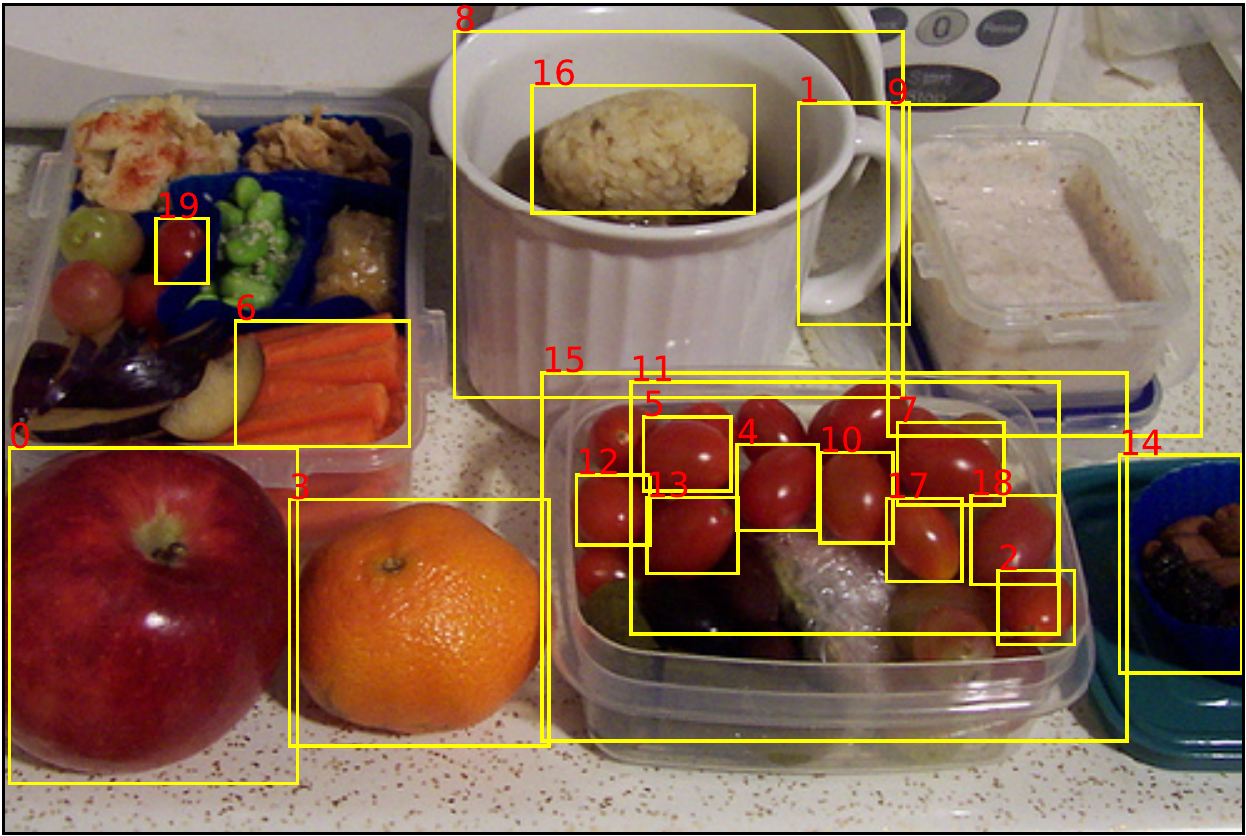}
	\includegraphics[width=0.69\linewidth]{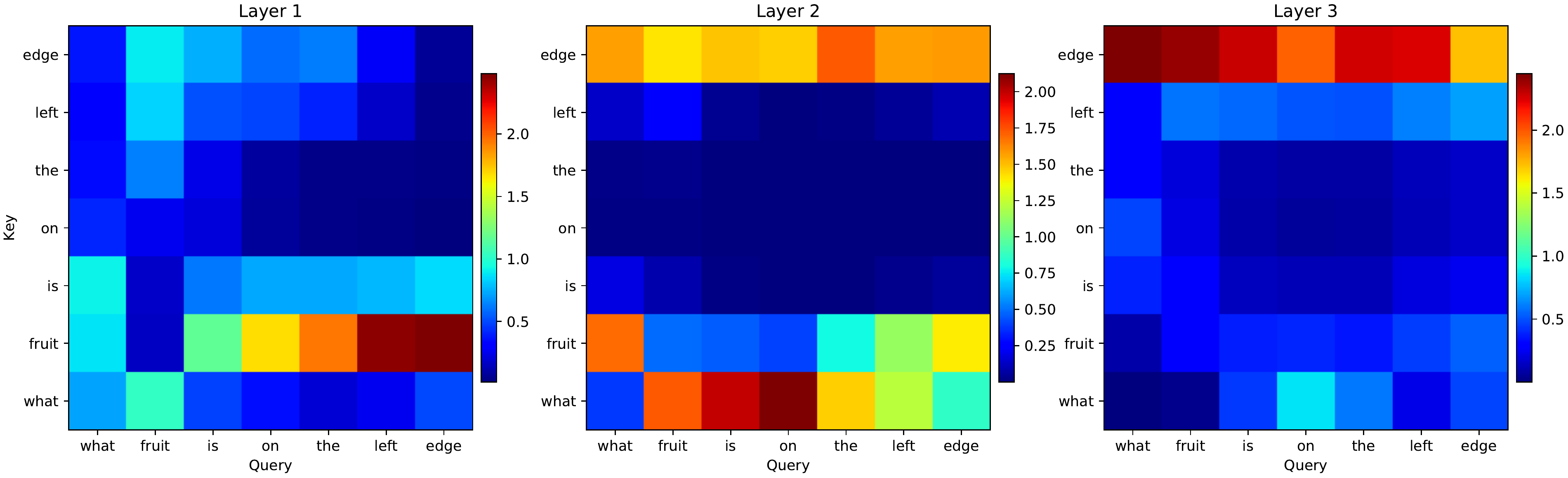}
	\includegraphics[width=1\linewidth, height=9cm]{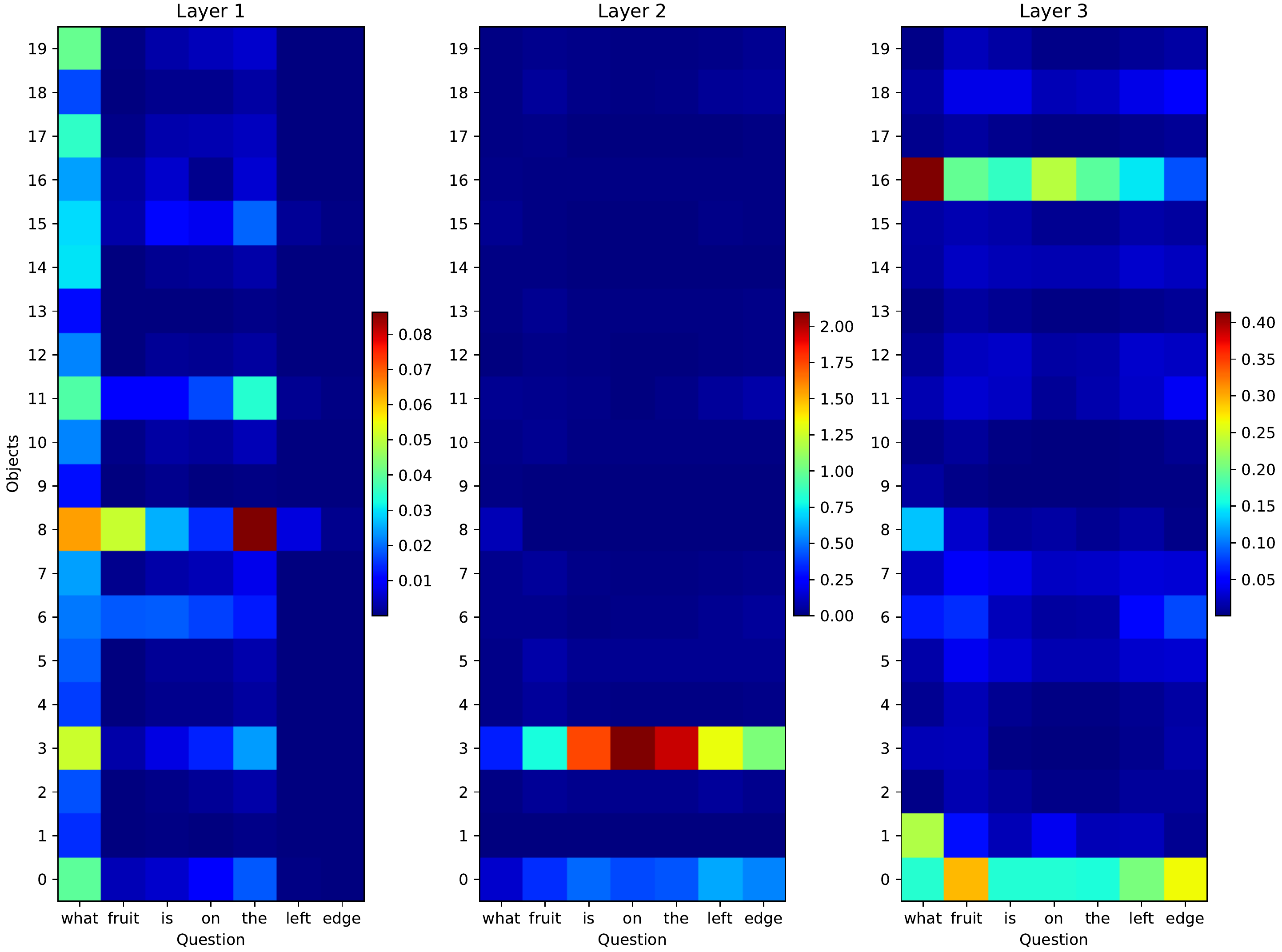}
	\caption{Visualization of attention maps for our networks. The attention maps in each graph for multiple glimpses are summed at each layer to briefly show the attended objects and words. The first image at the top shows the bounding boxes of detected objects and others for graph attention weights between words. The images at the bottom show the graph attention weights between words and objects. The predicted answers are tomato, orange, and apple respectively for one-layer, two-layer, and three-layer models of our bilinear graph networks.}
	\label{fig:attention_map}
\vspace{-5pt}
\end{figure*}

\subsection{Comparison with State-of-the-Art}
\begin{table}
	\centering	
	\resizebox{\linewidth}{!}{
	\begin{tabular}{lccccc}
		\hline
		\textbf{Model} & \textbf{Overall} & \textbf{Yes/no} & \textbf{Number} & \textbf{Other} & \textbf{Test-std} \\
		\hline
		Bottom-Up \cite{anderson2018bottom} & 65.32 & 81.82 & 44.21 & 56.05 & 65.67 \\
		Counter \cite{zhang2018learning} & 68.09 & 83.14 & 51.62 & 58.97 & 68.41 \\
		MuRel \cite{cadene2019murel} & 68.03 & 84.77 & 49.84 & 57.85 & 68.41 \\
		MFH+Bottom-Up \cite{yu2018beyond} & 68.76 & 84.27 & 49.56 & 59.89 & - \\
		BAN+Glove \cite{kim2018bilinear} & 69.66 & 85.46 & 50.66 & 60.60 & - \\
		BAN+Glove+Counter \cite{kim2018bilinear} & 70.04 & 85.42 & 54.04 & 60.52 & 70.35 \\
		DFAF \cite{gao2019dynamic} & 70.22 & 86.09 & 53.32 & 60.49 & 70.34 \\
		Vilbert \cite{lu2019vilbert} & 70.55 & - & - & - & 70.92 \\
		MCAN \cite{yu2019deep} & 70.63 & 86.82 & 53.26 & 60.72 & 70.90 \\
		MLIN \cite{gao2019multi} & 70.18 & 85.96 & 52.93 & 60.40 & 70.28 \\
		BGNs+Glove+Counter (ours) & 71.00 & 86.62 & \textbf{56.31} & 60.95 & - \\
		BGNs+Glove (ours) & 70.97 & 87.03 & 53.56 & 61.18 & 71.12 \\
		\hline
		DFAF+BERT \cite{gao2019dynamic} & 70.59 & 86.73 & 52.92 & 61.04 & 70.81 \\
		MLIN+BERT \cite{gao2019multi} & 71.09 & 87.07 & 53.39 & 60.49 & 71.27 \\
		BGNs+BERT (ours) & \textbf{72.28} & \textbf{88.60} & 54.09 & \textbf{62.46} & \textbf{72.41} \\
		\hline
	\end{tabular}}
	\vspace{0pt}
	\caption{Accuracy of single model on VQA v2.0 test-dev and test-standard dataset, it is trained on training, validation splits and Visual Genome dataset.}
	\label{table:vqa_model1}
	\vspace{-5pt}
\end{table}
In Table \ref{table:vqa_model1}, we evaluate our method on VQA v2.0 test-dev dataset, which achieves state-of-the-art. As shown in Table \ref{table:vqa_model1}, the overall accuracy of our BGNs+Glove model is 1.31\% higher than BAN+Glove, nearly 3.0\% on number metric. It can be explained that the counting task is a kind of relation among objects, which tries to find similar objects in the latter layers with objects grounded by previous layers. And the extra counter module \cite{zhang2018learning} in our BGNs+Glove+Counter model makes a little gain on overall accuracy since it might increase the counting ability but disturb our reasoning graphs leading to drop in other metrics. Thus, we choose BGNs+Glove and BGNs+BERT as our best models to evaluate them on the test-standard dataset. 

As we mentioned in Section \ref{sec:bgn}, our bilinear graph networks have a better explanation in modeling the relationship within multi-modal inputs, and we also achieve better performance on both test-dev and test-std dataset compared with other methods with and without BERT, proving the effectiveness of our proposed method.

\vspace{-5pt}
\subsection{Qualitative Analysis}
\begin{figure*}
	\centering
	\includegraphics[width=1\linewidth]{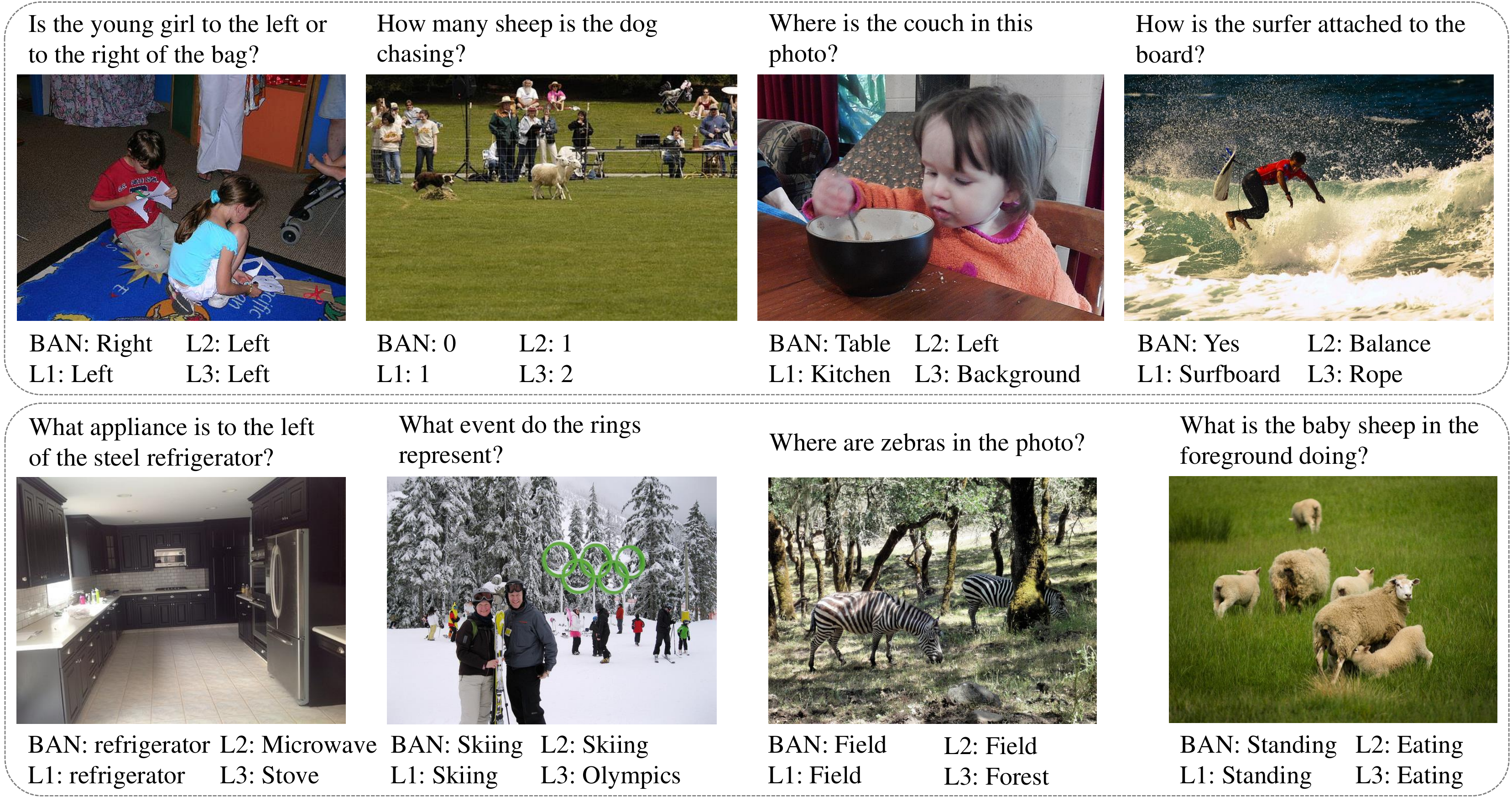}
	\caption{Examples illustrate the answers predicted by BAN and our graph models. BAN, L1, L2, L3 denote the answers predicted by BAN, one-layer, two-layer, and three-layer of our model respectively.}
	\label{fig:visual}
	\vspace{-5pt}
\end{figure*}
To visualize the effects of each module in our bilinear graph networks, we present the learned attention maps of the image-graph and the question-graph in each layer to show how the networks work. Given the question `What fruit is on the left edge?' in Figure \ref{fig:attention_map}, the image-graph of the first layer attends kinds of objects in the input image, while the question-graph broadcasts the learned fruit information to other words and chooses `tomato' as the answer, probably because the amount of `tomato' is the biggest among all detected fruits. The image-graph of the second-layer picks `orange' that is to the left of `tomato' and the question-graph keeps collecting `fruit' and `edge' information. In the third layer, the image-graph locates `apple' that is on the left edge, and every word in the question-graph pays its attention to the `edge' information to predict the answer.

In Figure \ref{fig:visual}, we show the answer predicted by BAN and our models with one layer, two layers, and three layers. In the first image of the top row, BAN cannot correctly answer the question because the entities of `young girl' and `bag' learn their positions respectively, but they do not know each other's information, while our proposed question-graph exchanges such positional information to make it possible to compare the relative direction of the two entities. A similar question can also be found in the first image of the bottom row, our model approaches the correct answer step by step as the layer of the graph increases. Moreover, our model can find the implicit relationship between objects, even when the sheep are far away from the dog in the second image of the top row, as well as abstract scenes in the second image (five circles representing Olympics) and third image (many trees composing forest) of the bottom row. Furthermore, our model finely discriminates the highly overlapped objects, such as two sheep in the second image and the rope in the fourth image of the top row, it is possibly because the question-graph undertakes some burden from the original graph of BAN, which makes the image-graph spare more effort on learning details in the image. 

\section{Conclusions}
Motivated by graph attention networks and Transformer, in this paper, we interpret bilinear attention networks from a new perspective and demonstrate its disadvantages, then we develop bilinear graph networks (BGNs) composed of layers of image-graph and question-graph to overcome them. The image-graph learns the graph between words in the question and objects in the image and generate the joint embeddings of them, while the question-graph models the graph between these joint embeddings in term of words to exchange context information. Our method achieves state-of-the-art performance on VQA v2.0 dataset, and the ablation studies show that our bilinear graph networks significantly outperform the BAN and other graph-based methods on a variety of questions.

{\small
	\bibliographystyle{ieee_fullname}
	\bibliography{2427}

\begin{thebibliography}{10}\itemsep=-1pt

\bibitem{anderson2018bottom}
Peter Anderson, Xiaodong He, Chris Buehler, Damien Teney, Mark Johnson, Stephen
  Gould, and Lei Zhang.
\newblock Bottom-up and top-down attention for image captioning and visual
  question answering.
\newblock In {\em CVPR}, volume~3, page~6, 2018.

\bibitem{antol2015vqa}
Stanislaw Antol, Aishwarya Agrawal, Jiasen Lu, Margaret Mitchell, Dhruv Batra,
  C Lawrence~Zitnick, and Devi Parikh.
\newblock Vqa: Visual question answering.
\newblock In {\em Proceedings of the IEEE international conference on computer
  vision}, pages 2425--2433, 2015.

\bibitem{cadene2019murel}
Remi Cadene, Hedi Ben-Younes, Matthieu Cord, and Nicolas Thome.
\newblock Murel: Multimodal relational reasoning for visual question answering.
\newblock In {\em Proceedings of the IEEE Conference on Computer Vision and
  Pattern Recognition}, pages 1989--1998, 2019.

\bibitem{chen2019graph}
Yunpeng Chen, Marcus Rohrbach, Zhicheng Yan, Yan Shuicheng, Jiashi Feng, and
  Yannis Kalantidis.
\newblock Graph-based global reasoning networks.
\newblock In {\em Proceedings of the IEEE Conference on Computer Vision and
  Pattern Recognition}, pages 433--442, 2019.

\bibitem{chung2014empirical}
Junyoung Chung, Caglar Gulcehre, KyungHyun Cho, and Yoshua Bengio.
\newblock Empirical evaluation of gated recurrent neural networks on sequence
  modeling.
\newblock {\em arXiv preprint arXiv:1412.3555}, 2014.

\bibitem{das2017visual}
Abhishek Das, Satwik Kottur, Khushi Gupta, Avi Singh, Deshraj Yadav,
  Jos{\'e}~MF Moura, Devi Parikh, and Dhruv Batra.
\newblock Visual dialog.
\newblock In {\em Proceedings of the IEEE Conference on Computer Vision and
  Pattern Recognition}, volume~2, 2017.

\bibitem{devlin2018bert}
Jacob Devlin, Ming-Wei Chang, Kenton Lee, and Kristina Toutanova.
\newblock Bert: Pre-training of deep bidirectional transformers for language
  understanding.
\newblock {\em arXiv preprint arXiv:1810.04805}, 2018.

\bibitem{fukui2016multimodal}
Akira Fukui, Dong~Huk Park, Daylen Yang, Anna Rohrbach, Trevor Darrell, and
  Marcus Rohrbach.
\newblock Multimodal compact bilinear pooling for visual question answering and
  visual grounding.
\newblock {\em arXiv preprint arXiv:1606.01847}, 2016.

\bibitem{gao2019dynamic}
Peng Gao, Zhengkai Jiang, Haoxuan You, Pan Lu, Steven~CH Hoi, Xiaogang Wang,
  and Hongsheng Li.
\newblock Dynamic fusion with intra-and inter-modality attention flow for
  visual question answering.
\newblock In {\em Proceedings of the IEEE Conference on Computer Vision and
  Pattern Recognition}, pages 6639--6648, 2019.

\bibitem{gao2019multi}
Peng Gao, Haoxuan You, Zhanpeng Zhang, Xiaogang Wang, and Hongsheng Li.
\newblock Multi-modality latent interaction network for visual question
  answering.
\newblock {\em arXiv preprint arXiv:1908.04289}, 2019.

\bibitem{goyal2017making}
Yash Goyal, Tejas Khot, Douglas Summers-Stay, Dhruv Batra, and Devi Parikh.
\newblock Making the v in vqa matter: Elevating the role of image understanding
  in visual question answering.
\newblock In {\em Proceedings of the IEEE Conference on Computer Vision and
  Pattern Recognition}, pages 6904--6913, 2017.

\bibitem{guo2019image}
Dalu Guo, Chang Xu, and Dacheng Tao.
\newblock Image-question-answer synergistic network for visual dialog.
\newblock In {\em Proceedings of the IEEE Conference on Computer Vision and
  Pattern Recognition}, pages 10434--10443, 2019.

\bibitem{hamilton2017inductive}
Will Hamilton, Zhitao Ying, and Jure Leskovec.
\newblock Inductive representation learning on large graphs.
\newblock In {\em Advances in Neural Information Processing Systems}, pages
  1024--1034, 2017.

\bibitem{he2016deep}
Kaiming He, Xiangyu Zhang, Shaoqing Ren, and Jian Sun.
\newblock Deep residual learning for image recognition.
\newblock In {\em Proceedings of the IEEE conference on computer vision and
  pattern recognition}, pages 770--778, 2016.

\bibitem{hochreiter1997long}
Sepp Hochreiter and J{\"u}rgen Schmidhuber.
\newblock Long short-term memory.
\newblock {\em Neural computation}, 9(8):1735--1780, 1997.

\bibitem{kim2018bilinear}
Jin-Hwa Kim, Jaehyun Jun, and Byoung-Tak Zhang.
\newblock Bilinear attention networks.
\newblock {\em arXiv preprint arXiv:1805.07932}, 2018.

\bibitem{kim2016hadamard}
Jin-Hwa Kim, Kyoung-Woon On, Woosang Lim, Jeonghee Kim, Jung-Woo Ha, and
  Byoung-Tak Zhang.
\newblock Hadamard product for low-rank bilinear pooling.
\newblock {\em arXiv preprint arXiv:1610.04325}, 2016.

\bibitem{kingma2014adam}
Diederik~P Kingma and Jimmy Ba.
\newblock Adam: A method for stochastic optimization.
\newblock {\em arXiv preprint arXiv:1412.6980}, 2014.

\bibitem{kipf2016semi}
Thomas~N Kipf and Max Welling.
\newblock Semi-supervised classification with graph convolutional networks.
\newblock {\em arXiv preprint arXiv:1609.02907}, 2016.

\bibitem{krishna2017visual}
Ranjay Krishna, Yuke Zhu, Oliver Groth, Justin Johnson, Kenji Hata, Joshua
  Kravitz, Stephanie Chen, Yannis Kalantidis, Li-Jia Li, David~A Shamma, et~al.
\newblock Visual genome: Connecting language and vision using crowdsourced
  dense image annotations.
\newblock {\em International Journal of Computer Vision}, 123(1):32--73, 2017.

\bibitem{lin2014microsoft}
Tsung-Yi Lin, Michael Maire, Serge Belongie, James Hays, Pietro Perona, Deva
  Ramanan, Piotr Doll{\'a}r, and C~Lawrence Zitnick.
\newblock Microsoft coco: Common objects in context.
\newblock In {\em European conference on computer vision}, pages 740--755.
  Springer, 2014.

\bibitem{liu2018structure}
Yong Liu, Ruiping Wang, Shiguang Shan, and Xilin Chen.
\newblock Structure inference net: Object detection using scene-level context
  and instance-level relationships.
\newblock In {\em Proceedings of the IEEE conference on computer vision and
  pattern recognition}, pages 6985--6994, 2018.

\bibitem{lu2019vilbert}
Jiasen Lu, Dhruv Batra, Devi Parikh, and Stefan Lee.
\newblock Vilbert: Pretraining task-agnostic visiolinguistic representations
  for vision-and-language tasks.
\newblock {\em arXiv preprint arXiv:1908.02265}, 2019.

\bibitem{nam2016dual}
Hyeonseob Nam, Jung-Woo Ha, and Jeonghee Kim.
\newblock Dual attention networks for multimodal reasoning and matching.
\newblock {\em arXiv preprint arXiv:1611.00471}, 2016.

\bibitem{pennington2014glove}
Jeffrey Pennington, Richard Socher, and Christopher~D. Manning.
\newblock Glove: Global vectors for word representation.
\newblock In {\em Empirical Methods in Natural Language Processing (EMNLP)},
  pages 1532--1543, 2014.

\bibitem{ren2015faster}
Shaoqing Ren, Kaiming He, Ross Girshick, and Jian Sun.
\newblock Faster r-cnn: Towards real-time object detection with region proposal
  networks.
\newblock In {\em Advances in neural information processing systems}, pages
  91--99, 2015.

\bibitem{salimans2016weight}
Tim Salimans and Durk~P Kingma.
\newblock Weight normalization: A simple reparameterization to accelerate
  training of deep neural networks.
\newblock In {\em Advances in Neural Information Processing Systems}, pages
  901--909, 2016.

\bibitem{srivastava2014dropout}
Nitish Srivastava, Geoffrey Hinton, Alex Krizhevsky, Ilya Sutskever, and Ruslan
  Salakhutdinov.
\newblock Dropout: a simple way to prevent neural networks from overfitting.
\newblock {\em The Journal of Machine Learning Research}, 15(1):1929--1958,
  2014.

\bibitem{vaswani2017attention}
Ashish Vaswani, Noam Shazeer, Niki Parmar, Jakob Uszkoreit, Llion Jones,
  Aidan~N Gomez, {\L}ukasz Kaiser, and Illia Polosukhin.
\newblock Attention is all you need.
\newblock In {\em Advances in Neural Information Processing Systems}, pages
  5998--6008, 2017.

\bibitem{velickovic2017graph}
Petar Velickovic, Guillem Cucurull, Arantxa Casanova, Adriana Romero, Pietro
  Lio, and Yoshua Bengio.
\newblock Graph attention networks.
\newblock {\em arXiv preprint arXiv:1710.10903}, 1(2), 2017.

\bibitem{wang2018non}
Xiaolong Wang, Ross Girshick, Abhinav Gupta, and Kaiming He.
\newblock Non-local neural networks.
\newblock In {\em Proceedings of the IEEE Conference on Computer Vision and
  Pattern Recognition}, pages 7794--7803, 2018.

\bibitem{yang2018graph}
Jianwei Yang, Jiasen Lu, Stefan Lee, Dhruv Batra, and Devi Parikh.
\newblock Graph r-cnn for scene graph generation.
\newblock In {\em Proceedings of the European Conference on Computer Vision
  (ECCV)}, pages 670--685, 2018.

\bibitem{yang2016stacked}
Zichao Yang, Xiaodong He, Jianfeng Gao, Li Deng, and Alex Smola.
\newblock Stacked attention networks for image question answering.
\newblock In {\em Proceedings of the IEEE Conference on Computer Vision and
  Pattern Recognition}, pages 21--29, 2016.

\bibitem{yu2019deep}
Zhou Yu, Jun Yu, Yuhao Cui, Dacheng Tao, and Qi Tian.
\newblock Deep modular co-attention networks for visual question answering.
\newblock In {\em Proceedings of the IEEE Conference on Computer Vision and
  Pattern Recognition}, pages 6281--6290, 2019.

\bibitem{yu2017mfb}
Zhou Yu, Jun Yu, Jianping Fan, and Dacheng Tao.
\newblock Multi-modal factorized bilinear pooling with co-attention learning
  for visual question answering.
\newblock {\em IEEE International Conference on Computer Vision (ICCV)}, pages
  1839--1848, 2017.

\bibitem{yu2018beyond}
Zhou Yu, Jun Yu, Chenchao Xiang, Jianping Fan, and Dacheng Tao.
\newblock Beyond bilinear: generalized multimodal factorized high-order pooling
  for visual question answering.
\newblock {\em IEEE transactions on neural networks and learning systems},
  (99):1--13, 2018.

\bibitem{yu2018rethinking}
Zhou Yu, Jun Yu, Chenchao Xiang, Zhou Zhao, Qi Tian, and Dacheng Tao.
\newblock Rethinking diversified and discriminative proposal generation for
  visual grounding.
\newblock {\em arXiv preprint arXiv:1805.03508}, 2018.

\bibitem{zhang2018learning}
Yan Zhang, Jonathon Hare, and Adam Pr{\"u}gel-Bennett.
\newblock Learning to count objects in natural images for visual question
  answering.
\newblock {\em arXiv preprint arXiv:1802.05766}, 2018.

\end{thebibliography}
}
\end{document}